\documentclass[11pt]{article}

\newif\ifanon \anonfalse

\IfFileExists{neurips_2026.sty}{%
  \usepackage[dblblindworkshop, final]{neurips_2026}
  \workshoptitle{AI for Stochastic Dynamics}
}{%
  \usepackage[margin=1in]{geometry}
  \PackageWarningNoLine{cosmos}{neurips_2026.sty not found -- fallback layout}
}

\usepackage[utf8]{inputenc}
\usepackage[T1]{fontenc}
\usepackage{hyperref}
\usepackage{url}
\usepackage{booktabs}
\usepackage{amsfonts}
\usepackage{amsmath}
\usepackage{amssymb}
\usepackage{microtype}
\usepackage{graphicx}
\usepackage{caption}
\usepackage{xcolor}

\newcommand{\vsM}[1]{#1}

\hypersetup{colorlinks=true,linkcolor=black,citecolor=black,urlcolor=blue}

\newcommand{\R}{\mbox{roll@20}}
\newcommand{\best}[1]{\textbf{#1}}

\title{COSMOS: Soft Mechanism Mixtures with Verifiable Routing for Long-Horizon PDE Forecasting}

\author{%
  Anupam Rawat \\
  Indian Institute of Technology Bombay \\
  \texttt{anupam.rawat@iitb.ac.in}
  \And
  Manikandan Padmanaban \\
  IBM Research \\
  \texttt{manipadm@in.ibm.com}
  \And
  Jagabondhu Hazra \\
  IBM Research \\
  \texttt{jahazra1@in.ibm.com}
}

\begin{document}
\maketitle

\begin{abstract}
Neural operators are an efficient alternative to classical PDE solvers, but most
learn one monolithic map per equation family and per discretization. Real
systems are more diverse: transport, diffusion, wave, and reaction processes
often act together. Existing mixture-of-experts operators follow the sparse
top-$K$ routing template, which selects a few experts and switches the rest off.
\vsM{When several processes act at once, the update is a blend, not a discrete
choice, so sparse selection throws away part of the physics.} Public benchmarks
such as PDEBench also evaluate equation families separately and carry no
mechanism-level mixture labels, so compositional routing cannot be tested on
them directly.

We propose COSMOS (Cooperative Operator Specialists with Mechanism-level
Operator Soft-routing), a soft mechanism-mixture neural operator. Four process-biased
specialists remain active at every step; a learned gate mixes them with
continuous weights, and a small network fuses their concatenated features.
Specialists share one coarse latent grid, and a parallel zero-initialized
full-resolution residual restores sharp detail the bottleneck discards. We also
introduce a compositional benchmark generated by operator splitting, with known
mixture weights $w^\star$ for every trajectory. On three PDEBench families, we
compare against the Fourier neural operator tuned per family (three learning
rates, five seeds each, best rate chosen on held-out validation) under a
20-step rollout protocol (normalized RMSE; lower is better): a locked 3-seed
run is ahead on diffusion--reaction ($0.241$ vs.\ $0.407$), roughly level on
Navier--Stokes ($0.493$ vs.\ $0.469$), and behind on shallow water ($0.091$
vs.\ $0.061$). Because a 3-seed mean can hide instability, we re-ran the same
recipe at 11 seeds: the Navier--Stokes result holds, but the
diffusion--reaction win does not -- that recipe diverges in a large minority
of seeds, and its 11-seed median trails FNO. We therefore do not claim a
reliable accuracy win on any of the three families; the audit itself, and
what it implies about trusting small-seed rollout comparisons, is as much
this paper's result as the architecture is. Retraining from scratch with the
gate fixed to uniform weights, or removing the specialist mixture while
keeping the residual, both degrade rollout substantially in the stable
regime. On the labeled benchmark, dense soft routing has $2.2\times$ lower
error than hard top-1 at identical fusion, and substituting the generator
coefficients $w^\star$ at inference lowers rollout to $0.034$. That gap is a
diagnostic of the learned gate, not a claim that $w^\star$ is optimal for the
trained specialists. The labels also falsify our own interpretability claim:
external routing does not align with the specialists' intended physics, so we
\vsM{claim accuracy, not mechanism identity.} We do not claim zero-shot
multi-family coverage, 3D, or unstructured meshes.
\end{abstract}

\section{Introduction}
\label{sec:intro}

Fast PDE surrogates matter wherever numerical simulation is the bottleneck,
from weather and subsurface flow to engineering design. Learned operators such
as the Fourier neural operator (FNO) \citep{li2021fno}, GINO \citep{li2023gino},
and Transolver \citep{wu2024transolver} make inference orders of magnitude
cheaper than classical solvers, but they share a structural assumption:
one map is fitted to one training distribution. A generalist PDE surrogate must
instead serve many mechanisms and their mixtures; COSMOS (Cooperative Operator Specialists with Mechanism-level Operator
Soft-routing) is designed as a step in that direction. Mixture-of-experts operators are
the natural response to physical diversity
\citep{moepot2025,nestor2026,imooe2025,compno2026}, and most adopt the sparse
template from large-scale language modeling: a gate selects a top-$K$ expert and
switches the rest off.

\vsM{This is a mismatch of kind, not of how many experts to keep.} A state that is
partly advective and partly diffusive needs both processes at once; a hard
gate must drop one of them, and that choice is least reliable exactly where
the two processes overlap. The gate should stay \emph{dense}: keep every
specialist active and let continuous weights carry the blend.
Section~\ref{sec:compositional} measures the cost: at identical fusion, hard
top-1 routing has $2.2\times$ higher rollout error than soft, \vsM{against the
sparse template most MoE operators take from language modeling.} Externally,
against a properly tuned FNO, a locked 3-seed run is roughly level on one of
three PDEBench families and ahead on a second, but a follow-up seed audit
shows the second win is not reliable (Section~\ref{sec:external}).

\paragraph{A verification problem sits underneath the modeling problem.}
MoE operators are routinely described as if experts \emph{specialize} and
routing \emph{discovers} structure. These are claims about what a model has
learned, and on standard benchmarks they are
\vsM{unfalsifiable}: PDEBench
\citep{takamoto2022pdebench} and comparable corpora are organized by equation
family and carry no trajectory-level record of which processes were active in
what proportion. A model can be called mechanism-aware with no experiment able
to show otherwise.

This paper treats that gap as the methodological contribution rather than a
caveat. We construct a corpus by operator splitting in which every trajectory
records the mixture $w^\star$ that generated it, supplying two things no
public benchmark can: a \emph{generator-truth substitution} that measures what
changes when the learned gate is replaced by the known coefficients, and a
\emph{falsification test} for the mechanistic reading of the router. Applied
to our own model, the test shows that external routing does not align with
the specialists' intended physics;
\vsM{we claim accuracy rather than mechanism identity}, and take the result as a working demonstration of the instrument.

\paragraph{Contributions.}
\begin{enumerate}\itemsep2pt
\item \textbf{Method.} COSMOS: dense continuous gating over process-biased
specialists, concat$+$MLP fusion, a shared latent library, and a zero-initialized
full-resolution residual (Section~\ref{sec:method}).
\item \textbf{Instrument.} A compositional corpus with ground-truth mixture
labels, enabling a generator-truth routing diagnostic and a controlled
routing ablation that public benchmarks cannot support
(Section~\ref{sec:benchmark}).
\item \textbf{Evidence, controls, and a self-audit.} Shared-harness PDEBench
results against a per-family tuned FNO, three ablative controls, a
soft-vs-hard comparison at fixed fusion, and explicit negative results,
including an 11-seed audit that overturns the locked 3-seed DR win
(Sections~\ref{sec:external}--\ref{sec:limits}).
\end{enumerate}

\paragraph{Scope.}
A small-scale, architecture-first study: COSMOS $\sim$3.4M parameters vs.\ FNO
$\sim$1.2M, a $16{\times}16$ latent, grid-only external tables; synthetic and
PDEBench models are trained separately and never averaged. All experiments use
deterministic solvers; we do not evaluate stochastic forcing or calibrated
uncertainty.

\section{Related Work}
\label{sec:related}

\paragraph{Neural operators.}
FNO \citep{li2021fno} is the standard grid baseline and our primary comparison.
GINO \citep{li2023gino} and Transolver \citep{wu2024transolver} extend operator
learning beyond regular grids; message-passing solvers
\citep{brandstetter2022message} emphasize local updates and rollout-aware
training, whose unrolled objective we adopt. As in Perceiver~IO
\citep{jaegle2022perceiver}, we map variable-resolution inputs onto a fixed latent
grid. A separate line decomposes dynamics by term or prior: PDE-Net learns
per-term differential operators \citep{long2018pdenet}, APHYNITY splits physics
from a learned residual \citep{yin2021aphynity}, PhyDNet uses a two-stream
physics$+$residual design \citep{guen2020phydnet}. COSMOS differs in routing a
\emph{library} of process operators with learned continuous weights, evaluated
against generator-known mixture labels on a controlled compositional
benchmark. The labels are used for evaluation, not as a training signal.

\paragraph{Mixture and compositional operators.}
CompNO \citep{compno2026} assembles pretrained process blocks; iMOOE
\citep{imooe2025} uses an invariance-aligned mixture of operator experts; MoE-POT
\citep{moepot2025} and NESTOR \citep{nestor2026} scale sparse MoE routing for
multi-PDE pretraining. COSMOS differs on three axes: dense soft gating rather
than top-$K$ sparsity (a return to the classical mixture formulation of
\citet{jacobs1991moe}; top-$K$ sparsity exists to save compute at hundreds of
experts, and is
\vsM{the wrong default at small $M$}), concat$+$MLP fusion rather than a weighted
sum, and process-biased specialists rather than homogeneous blocks. Our
comparisons against MoE-POT and NESTOR are matched \emph{from-scratch}
contact tests at small scale, outside the pretraining regime those methods
target.

\paragraph{Verification and interpretability of learned operators.}
The claim that an expert has specialized is an interpretability claim, and
plausible attributions often fail under controlled test. Our contribution is
narrow: a labeled generator makes the claim checkable, and we report a case
where it does not survive. A companion study \citep{spectralprobe2026} finds
gated multi-stream encoding outperforms flat concatenation for PDE-family
identification; we adopt the same principle.

\section{Method}
\label{sec:method}

\begin{figure}[t]
\centering
\includegraphics[width=0.68\textwidth]{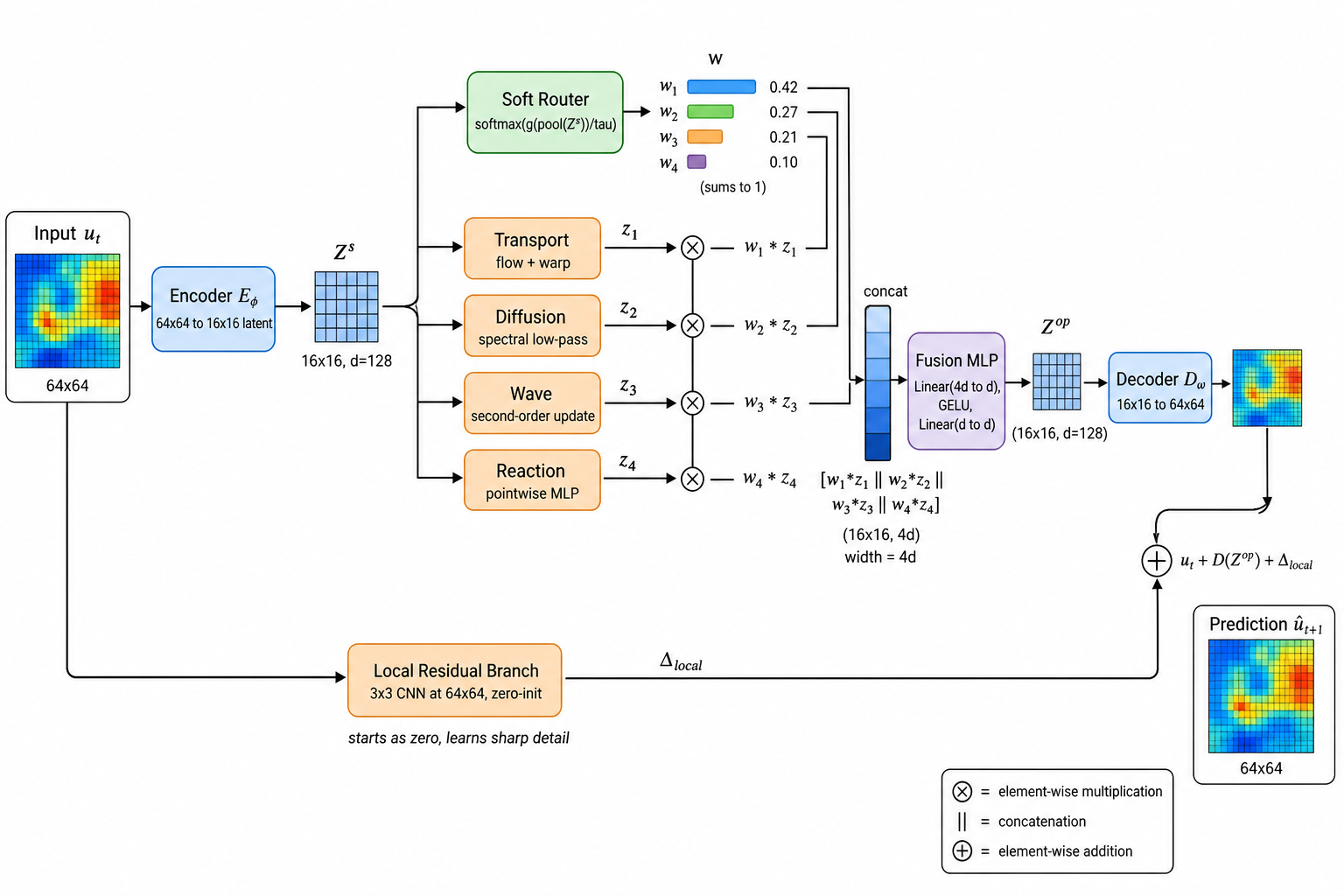}
\caption{COSMOS forward pass. Frames are encoded onto a shared $16{\times}16$
latent, processed in parallel by four process-biased specialists, mixed by a
\emph{dense} router (no top-$K$), fused by concatenation$+$MLP, and decoded. A parallel full-resolution residual is zero-initialized, so training begins
exactly at the latent-only update and the branch
\vsM{learns only what the coarse latent cannot represent}. (Figure notation: $Z^{\mathrm{op}}=\hat z^t$,
$\Delta_{\mathrm{local}}=R_\rho$; the decoder predicts the change from $u_t$.)}
\label{fig:arch}
\end{figure}

We target dynamics expressible as a combination of process operators,
\begin{equation}
\frac{\partial u}{\partial t}=\sum_{m=1}^{M}\alpha_m(x,t)\,\mathcal{L}_m[u]+\varepsilon,
\qquad M=4,
\label{eq:decomp}
\end{equation}
with transport, diffusion, wave, and reaction in the library. The term
$\varepsilon$ collects unmodeled physics, discretization error, and any
stochastic forcing. Every dataset here comes from a deterministic solver, so
we set $\varepsilon=0$; we do not evaluate noise or calibrated uncertainty.

The surrogate learns $u^{t+1}=\mathcal{G}_\theta(u^t,\dots,u^{t-k+1})$ and is
applied autoregressively; mixture structure must be inferred from the encoded
input history, without access to $w^\star$.

\subsection{Canonical latent and process-biased specialists}
The encoder $E_\phi$ maps $k$ past frames at native resolution to
$z^t\in\mathbb{R}^{d\times H_c\times W_c}$ with $H_c{=}W_c{=}16$, $d{=}128$. The
specialist library is defined once on this fixed grid, so only $E_\phi$ and the
decoder $D_\omega$ change for a different input layout. Each specialist $f_m$ carries a lightweight architectural bias: transport
predicts a displacement field and warps the latent by it; diffusion applies a
learned per-wavenumber low-pass filter; wave applies a spectral filter with a
learned dispersion phase (a bounded per-wavenumber complex rotation); reaction is
a pointwise MLP with no spatial coupling. These biases shape
the hypothesis space; they do not enforce semantic mechanism identity after
training, which is precisely the gap Section~\ref{sec:routing} measures.

\subsection{Dense routing and concatenation fusion}
A router reads a pooled summary of $z^t$ and emits continuous weights
\begin{equation}
w=\mathrm{softmax}\!\left(g_\psi(\mathrm{mean\text{-}pool}(z^t))/\tau\right)\in\Delta^{M-1},
\label{eq:router}
\end{equation}
Every specialist is evaluated at every step; there is no top-$K$ selection.
Rather than forming $\sum_m w_m f_m(z^t)$, we concatenate weighted specialist
features and fuse,
\begin{equation}
\hat z^t=\mathrm{MLP}_\xi\!\left(\big[\,w_1f_1(z^t)\,\Vert\cdots\Vert\,w_Mf_M(z^t)\,\big]\right),
\label{eq:fusion}
\end{equation}
which can represent interactions between specialist outputs, not only their convex
combination. Both choices are ablated in Section~\ref{sec:compositional}: the
weighted sum costs $0.176$ against concatenation's $0.137$, and hard top-1 costs
$0.301$.

\subsection{Full-resolution residual}
\label{sec:bridge}
Compressing a $64{\times}64$ field onto a $16{\times}16$ latent
\vsM{discards sharp detail}. We add a native-resolution correction,
\begin{equation}
u^{t+1}=u^{t}+D_\omega(\hat z^t)+R_\rho(u^t,\dots,u^{t-k+1}),
\label{eq:residual}
\end{equation}
with the final layer of $R_\rho$ zero-initialized, so optimization begins
exactly at the latent-only update $u^{t}+D_\omega(\hat z^t)$ and the mixture is
undisturbed at step $0$.

\subsection{Objective}
We minimize an unrolled multi-step loss with mildly increasing step weights
$\lambda_k$,
\begin{equation}
\mathcal{L}=\sum_{j=1}^{K}\lambda_j \frac{\big\Vert\hat u^{\,t+j}-u^{t+j}\big\Vert_2}{\big\Vert u^{t+j}\big\Vert_2} +\beta\,\mathcal{L}_{\mathrm{route}},
\label{eq:loss}
\end{equation}
where $\mathcal{L}_{\mathrm{route}}$ adds entropy and load-balancing
regularization on $w$, discouraging premature collapse onto a single
specialist.

\section{An Instrument for Mechanism Claims}
\label{sec:benchmark}

External benchmarks establish whether a design transfers. They cannot establish
\emph{why} it works, because they carry no mechanism labels. We therefore generate
a corpus by operator splitting, integrating each process operator in alternation with known coefficients, so the generator records exactly which processes acted.

\paragraph{Construction.}
Each trajectory carries $w^\star=(w_T,w_D,w_W,w_R)$ summing to one, derived from
the generator coefficients. Four \emph{pure} families activate a single mechanism;
five \emph{mixed} families compose two or three: advection--diffusion
(transport$+$diffusion; ``advection'' and ``transport'' are used
interchangeably), reaction--diffusion, damped wave (wave$+$diffusion),
advection--reaction--diffusion, and damped-wave--reaction (wave$+$diffusion$+$reaction). After integration on a
$33{\times}33$ grid, each trajectory is resampled onto an unstructured mesh and a
point cloud ($n{=}1024$), sharing $w^\star$ and coefficients and differing only in
coordinate layout. These are \textbf{constructed siblings}, not independent
unstructured simulations, and we use them only as a transfer check.

\paragraph{Splits.}
The primary split \textbf{S2} trains on all four pure families (grid$+$mesh) and
tests on unseen three-mechanism mixtures, a compositional generalization test
within the mechanisms the generator represents.
The harder split \textbf{S1} trains on pure transport and diffusion only and tests
on adv-diff, a deliberately out-of-regime probe.

\paragraph{What the labels buy.}
$w^\star$ is never used as a training signal for any model reported here; it is
the normalized relative strengths of the generator's per-process coefficients,
and serves evaluation only. It supplies two things that make a mechanism claim
testable. \textbf{A generator-truth substitution:} replacing the learned
weights by $w^\star$ at inference, specialists unchanged, measures how much
performance depends on the gate. The resulting gap is a diagnostic of the
learned router; it does not assume that $w^\star$ is the optimal routing
vector for the trained specialists. \textbf{A falsification test:}
correlating learned $w$ against $w^\star$ asks whether the router recovers
the generator's mixture; externally, where $w^\star$ does not exist, we can
only ask the weaker qualitative question of whether routing mass aligns with
each family's known physics (Section~\ref{sec:routing}). $w^\star$ exists
because we generated the data and cannot be recovered post hoc for PDEBench,
so the corpus licenses a mechanism claim only where the generating mixture
is known.

\section{Results}
\label{sec:external}

\paragraph{Protocol.}
All headline numbers are \R{}: RMSE after 20 autoregressive steps, divided by
the norm of the true field, so $0$ is perfect and the trivial repeat-last-frame
baseline (\emph{persistence}) sits at $2.51$ on shallow water.
One-step error is a secondary diagnostic: an order of magnitude smaller than
rollout error and closely clustered across methods, so it does not distinguish
them. COSMOS ($\sim$3.4M, $d{=}128$) and FNO ($\sim$1.2M) are evaluated on
$64{\times}64$ PDEBench grids under one shared rollout harness. COSMOS uses
three seeds, the setting under which its recipe was locked. For FNO, since a
single learning rate does not fit every family well, we trained it at three
rates and five seeds per rate, picked the best rate on held-out validation,
and report only that rate's five test seeds -- a broader hyperparameter
search than COSMOS received, so the comparison is not driven by an
obviously suboptimal FNO learning rate.

\paragraph{The latent bottleneck is real, and it motivates the residual.}
Before the residual is enabled, latent-only COSMOS on diffusion--reaction (DR)
sits essentially at persistence (the trivial baseline that repeats the last input frame). Roughly $95\%$ of the initial-condition spectral
energy on these trajectories lies above the $16{\times}16$ latent Nyquist, so the
coarse library \vsM{cannot carry the sharp modes at all}. This is a statement about
these initial conditions at this latent resolution, not a general claim about
latent operators. It is, however, what Equation~\eqref{eq:residual} exists to fix, and
every number below includes that path.

\begin{table}[t]
\centering
\caption{PDEBench \R{} under the shared evaluation harness, lower is better.
COSMOS: mean$\pm$std over 3 seeds ($\sim$3.4M parameters). FNO: mean$\pm$std
over 5 seeds at its best learning rate, chosen per family on a held-out
validation split ($\sim$1.2M parameters). This is not a strict capacity
match; three controls reported in the text address different sources of the
observed differences.}
\label{tab:ext}
\begin{tabular}{lccc}
\toprule
Family & COSMOS & FNO \citep{li2021fno} & Ratio \\
\midrule
Diffusion--reaction (DR) & $0.241{\pm}0.061^\dagger$ & $0.407{\pm}0.211$ & $0.59\times^\dagger$ \\
Navier--Stokes (NS2D)    & $0.493{\pm}0.078^\dagger$ & \best{$0.469{\pm}0.024$} & $1.05\times^\dagger$ \\
Shallow water (SWE)      & $0.091{\pm}0.030$ & \best{$0.061{\pm}0.029$} & $1.49\times$ \\
\bottomrule
\end{tabular}\\[2pt]
{\footnotesize $\dagger$ Locked 3-seed protocol estimate, not the recipe's
typical seed: \S\ref{par:control3}'s 11-seed audit gives DR2D median $0.548$
(worse than FNO) and NS2D median $0.469$ (matches).}
\end{table}

\paragraph{Headline.}
Table~\ref{tab:ext} reports the comparison, with FNO tuned as described above
rather than run at one fixed rate: the single rate we tried first was a poor
fit for two of three families, leaving FNO $2$--$13\times$ worse than its own
best rate on DR and NS2D, while already close to best on SWE. At the locked
3-seed protocol, COSMOS leads on DR, is roughly level on NS2D, and trails on
SWE. The table footnote gives the reason we do not stop there: \S\ref{par:control3}'s
11-seed audit finds DR's outcome is bimodal and its median seed loses to
FNO, so we do not claim a reliable DR2D win; NS2D's locked number matches its
11-seed median almost exactly, so that result stands. We report the tuned
baseline, weaker for COSMOS on two of three families, because it is the fair
comparison, and we report the audit alongside the locked table because the
locked number alone would overstate DR2D.

\paragraph{Control 1: is the gate doing anything?}
Training the full recipe from scratch with the gate fixed to uniform weights
throughout, rather than learned, raises DR error from $0.241$ to $2.54$, a
collapse past persistence. This is a from-scratch retrain, not a
swap on an already-trained model: fixing the gate before training changes
what the specialists themselves learn to do: every specialist then receives
the same fixed weight rather than an input-dependent training signal. A
learnable dense gate is therefore part of the training recipe, not a
post-hoc mixer; \S\ref{sec:routing} returns to what the \emph{learned} gate
does once training is finished.

\paragraph{Control 2: is the residual doing all the work?}
Because FNO carries no comparable native-resolution branch, we isolate the
residual's contribution directly. A residual-only model that removes the mixture
entirely but keeps the full-resolution path reaches $0.295$ on DR (seed0;
three-seed mean $0.516{\pm}0.472$, inflated by one divergent seed) against
the full model's $0.241$. The mixture therefore improves on the residual alone, and on
the well-behaved seeds the residual-only model still sits below FNO's tuned
mean of $0.407$.
\vsM{Both ingredients contribute; neither explains the result by itself.}

\paragraph{Control 3: are the process-biased specialists themselves reliable?}
\label{par:control3}
A 3-seed mean can look stable even when individual seeds are not (\S\ref{sec:external}), so we trained the same locked recipe at 11 random seeds on DR
and on NS2D, together with a control in which the four process-biased
specialists are replaced by four generic blocks of equal size and no physics
bias. The generic control uses 11 seeds on DR and the eight completed seeds
on NS2D. We call a rollout error above $1.0$, worse than persistence, a
\emph{divergence}.

On DR, the process-biased specialists diverge on $4$ of $11$ seeds. The
worst of these is $34.8\times$ higher error than the good seeds. The generic
blocks diverge on only $1$ of $11$ seeds, and that one case is barely over
the threshold, not a collapse. On NS2D the pattern reverses: the
process-biased specialists diverge on $2$ of $11$ seeds, while the generic
blocks diverge on $3$ of $8$.

In every seed, good or bad, the one-step error looks the same: normal and
small. The difference only shows up in the rollout. Error tracks normally
through roughly step $10$, then grows sharply by step $20$. So the failure
sits in the autoregressive loop, not in training. It is also not caused by
the routing weights: forcing the learned gate to exactly uniform weights at
inference, on both a stable and a divergent seed, changes the result by
under $0.01\%$. The immediate failure is therefore not the precise
inference-time gate; it sits in the learned specialist and fusion dynamics.
That does not rule out a role for routing during training.

We read this as two separate findings, not one. Process-biased specialists
carry a real robustness cost on DR specifically, and the 11-seed median
($0.548$) trails FNO's tuned mean ($0.407$): the locked 3-seed win does not
hold at typical seeds. NS2D, on the other hand, is rollout-fragile for every
method we tested there, FNO included, independent of architecture, and its
11-seed median ($0.469$) matches the locked estimate, so that result stands.

\paragraph{Contact with MoE operators.}
At matched $\sim$4M from-scratch capacity, COSMOS reaches $0.28$ on DR and
$0.137$ on synthetic S2, against NESTOR ($1.12$, $0.452$) and MoE-POT ($2.21$,
$1.074$); a time-boxed fine-tune of a pretrained MoE-POT Tiny checkpoint on DR
reached $0.928$. These are contact probes at matched small scale, not evidence
against either method in its intended multi-PDE-pretrained regime.

\section{Routing Analysis on the Labeled Benchmark}
\label{sec:compositional}

\begin{table}[t]
\centering
\caption{Compositional S2 holdout, grid, \R{} (3 seeds). All COSMOS rows use
identical concat$+$MLP fusion, so the routing comparison is clean.
Generator-truth substitutes $w^\star$ at inference while leaving the trained
specialists unchanged: a diagnostic, not a baseline or a guaranteed
optimum. ``---'' = not recorded under this protocol.}
\label{tab:synth}
\begin{tabular}{lcc}
\toprule
Method & one-step & \R{} \\
\midrule
\emph{Generator-truth $w^\star$ (diagnostic)} & $0.0021{\pm}0.0003$ & \emph{$0.034{\pm}0.013$} \\
\midrule
COSMOS dense soft routing (primary)  & $0.0031{\pm}0.0001$ & \best{$0.137{\pm}0.065$} \\
\quad weighted-sum fusion   & $0.0032{\pm}0.0001$ & $0.176{\pm}0.023$ \\
\quad soft hierarchical     & $0.0032{\pm}0.0001$ & $0.220{\pm}0.057$ \\
\quad uniform weights       & $0.0032{\pm}0.0001$ & $0.283{\pm}0.056$ \\
\quad hard top-1            & --- & $0.301{\pm}0.006$ \\
\midrule
FNO \citep{li2021fno}       & $0.0034{\pm}0.0001$ & $0.326{\pm}0.038$ \\
GINO \citep{li2023gino} ($2.7$M) & --- & $0.369{\pm}0.021$ \\
Persistence                 & $0.0035$ & $0.370$ \\
NESTOR \citep{nestor2026}   & $0.0046{\pm}0.0005$ & $0.452{\pm}0.233$ \\
Dense monolith (matched)    & $0.0236{\pm}0.0109$ & $0.572{\pm}0.162$ \\
MoE-POT \citep{moepot2025}  & $0.0065{\pm}0.0014$ & $1.074{\pm}0.592$ \\
\bottomrule
\end{tabular}
\end{table}

\paragraph{Dense soft routing versus hard top-1.}
Table~\ref{tab:synth} holds fusion fixed and varies only the gate, which makes the
comparison attributable. Soft-flat routing ($0.137$) improves on hierarchical
routing ($0.220$), uniform weights ($0.283$), and hard top-1 ($0.301$). The last
comparison is the one that carries the argument of
Section~\ref{sec:intro}: at matched fusion, dense continuous routing is
$2.2\times$ better than \vsM{the sparse template MoE operators inherit from language
modeling}. Paired per-trajectory Wilcoxon tests rank the primary configuration
above every learned baseline at $p<10^{-40}$
(Appendix~\ref{app:wilcoxon}); those tests are conditional on the trained
checkpoints and do not replace seed-level uncertainty.

\paragraph{\vsM{Elaboration does not help.}}
Hierarchical routing, cross-expert attention, shared experts, and gated
constraints were all implemented; none improved the chain in either fusion mode
(Appendix~\ref{app:ablation}). We report the compact configuration that
survives, not the one we expected.

\paragraph{Where the error actually lives.}
Replacing the learned router by $w^\star$ lowers \R{} from $0.137$ to
$0.034$, about $4\times$, with the specialists left unchanged. Routing is
therefore a large, recoverable source of error in this controlled setting.
That does not make $w^\star$ the optimal routing vector for the trained
specialists; it is a diagnostic that public benchmarks cannot supply.

\begin{figure}[t]
\centering
\includegraphics[width=0.55\textwidth]{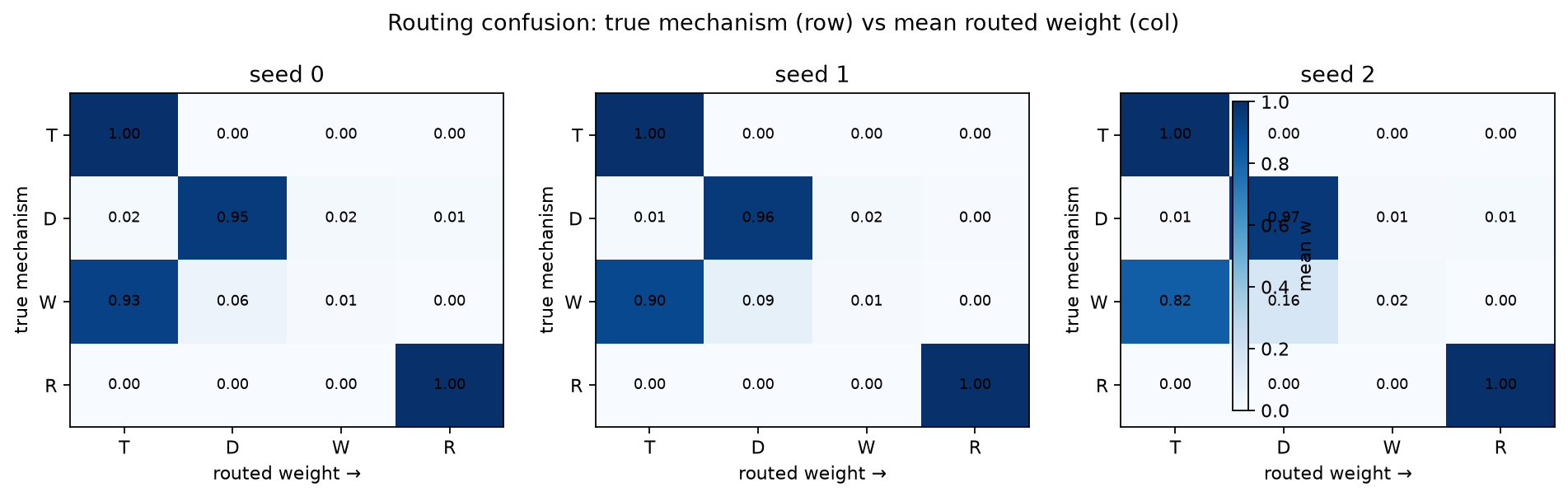}
\caption{Mean routed weight per specialist on pure-mechanism trajectories,
three seeds. The largest mean weight aligns with the correspondingly labeled
specialist for three of four mechanisms; pure wave routes to transport
($0.82$--$0.93$), a spectral ambiguity discussed in
Section~\ref{sec:routing}.}
\label{fig:routing}
\end{figure}

\paragraph{Routing diagnostics.}
\label{sec:routing}
On the compositional corpus, learned weights correlate with $w^\star$ at
$0.67$--$0.69$ across seeds, and mean routed weight lands on the
correspondingly labeled specialist for three of the four mechanisms
(transport $1.00$, diffusion $0.95$--$0.97$, reaction $1.00$ across seeds;
Figure~\ref{fig:routing}). The fourth is instructive: pure-wave trajectories
route to the \emph{transport} specialist. Linear and nonlinear probes did not
support pooling loss, router capacity, or data diversity as sufficient
explanations: transport and wave are both spectrum-\emph{preserving}
operators, so a single frame of either is statistically a band-limited field
(encoded separability caps at ${\approx}0.76$, vs.\ $0.96$--$1.00$ for every
other pair). The distinction lives in temporal evolution (first-order
directional vs.\ second-order isotropic), which the pooled router summary does
not see; diffusion and reaction \emph{transform} the spectrum and are
identifiable from one frame. Mixture accuracy is unaffected, since soft fusion
keeps both operators active -- itself an argument for dense routing over a hard
choice.

On external PDEBench diffusion--reaction, routing mass concentrates on the
transport and wave specialists rather than diffusion and reaction. We therefore
do not claim mechanism identity on external data, and read the specialist names
as design-time inductive biases throughout.

\paragraph{What the trained gate actually does at inference.}
Control 1 shows a gate fixed to uniform \emph{from the start of training}
costs accuracy -- a claim about training, not about the \emph{converged,
learned} gate, so we checked the latter directly. Over $5000$ external DR
predictions, the learned gate's per-sample entropy sits at $98\%$ of its
maximum: on almost every input it outputs close to equal weight on all four
specialists. Forcing a trained checkpoint's weights to exactly uniform
\emph{at inference}, no retraining, changes the result by under $0.01\%$ on
both a well-behaved and a divergent Control~3 seed. Turning off the
training-time regularizer that discourages a sharp gate has the opposite
effect: routing sharpens (entropy falls to $42\%$ of maximum) and rollout
error nearly quadruples. That last observation is a single seed, so we treat
it as diagnostic rather than conclusive.

Put together: the regularizer appears to keep the gate from collapsing early,
which would starve the other specialists of gradient. For the external DR
checkpoints we inspected, the precise converged weights matter little at
inference: replacing their near-uniform outputs by exact uniforms changes
rollout by under $0.01\%$. Control 1 still stands --- a learnable gate
matters to the training recipe --- but the external DR evidence does not
support reading the converged gate as selecting a physically correct mixture
on each input.

\paragraph{Constructed siblings.}
Reusing the same trained weights on mesh and point-cloud siblings degrades \R{}
from $0.137$ to $0.175$ ($1.27\times$) and $0.180$ ($1.31\times$), supporting a
single latent library across layouts on \emph{constructed} siblings only --
not evidence on independent unstructured solvers.

\paragraph{Scaling, preliminary.} A capacity-matched joint pretrain over five 2D
families, followed by a short per-family fine-tune, recovers rollout at or
below the specialist solo on all five, and the same recipe transfers to five
1D families (Appendix~\ref{sec:scale}). These runs lack the seed-count audit
of \S\ref{par:control3}, so we report them as preliminary only.

\section{Limitations}
\label{sec:limits}

\textit{Mechanism identity.} The routing diagnostics of Section~\ref{sec:routing}
hold on the labeled corpus, not externally: on PDEBench DR, routing mass does
not align with the specialists' intended physics. We claim no
interpretability on external data; specialist names are design-time biases.

\textit{Robustness, and this paper's clearest limitation.} \S\ref{par:control3}'s
11-seed audit overturns the DR2D win, not just caveats it: the median seed
trails FNO, and the process-biased specialists diverge far more often than a
matched generic-expert control on DR. We could not identify which specialist
causes it, or a fix, in the time available; reporting this over the locked
3-seed table matters more than a result that looks cleaner than the method
is. Three-seed comparisons are underpowered for training-run variability;
Appendix~\ref{app:wilcoxon} reports paired trajectory tests separately and
does not mix them with seed-level statistics.

\textit{Out-of-regime mixtures.} On the harder S1 split (train on two
mechanisms, test on their unseen combination), COSMOS trails FNO: soft
mixtures do not extrapolate beyond the trained regime.

\textit{Baseline matching and scale.} COSMOS carries more parameters than FNO
and a native-resolution branch FNO lacks (the residual-only control bounds
this rather than eliminating it), and trains multi-step unrolled where FNO
trains one-step; whether a rollout-trained FNO narrows the gap is open.
Increasing the external model's width also degrades DR rollout, unexplained
so far; external tables are regular grids only, mesh/point-cloud results are
constructed siblings not independent simulations, single-seed 3D results
trail specialized baselines, and MoE-operator comparisons are from-scratch
at matched capacity, outside those methods' pretraining regime.

\section{Conclusion}

COSMOS mixes process-biased specialists with a dense continuous gate,
concat$+$MLP fusion, and a zero-initialized full-resolution residual. A
locked 3-seed run is \vsM{ahead of tuned FNO on DR and level on NS2D}; the 11-seed
audit overturns the DR result at typical seeds (\S\ref{par:control3}) and
leaves the NS2D estimate unchanged. A labeled generator and a larger seed
sweep each falsified one of our own claims. Closing the
generator-truth--learned routing gap and isolating DR's unstable specialist
are next.

\bibliographystyle{plainnat}
\bibliography{refs}

\appendix
\newpage

\section{Scaling the Recipe}
\label{sec:scale}

\paragraph{Five 2D families with joint pretraining.}
Beyond per-family specialists, we train one capacity-matched joint backbone
($d{=}256$, $K_m{=}8$, $\sim$16.2M) on five PDEBench 2D families and apply a short
per-family fine-tune. Zero-shot joint transfer is \emph{not} sufficient (SWE in
particular degrades sharply), but after fine-tuning, rollout recovers at or below
the specialist solo on all five families (Table~\ref{tab:stage2}). We do not claim
a single frozen checkpoint for all of PDEBench, and the capacity here differs from
Table~\ref{tab:ext}, so rows are not comparable across the two.

\begin{table}[h]
\centering
\caption{Joint pretrain then short per-family fine-tune, \R{}. Ratio is
fine-tuned/solo (acceptance gate $\leq1.10\times$; 5/5 pass).}
\label{tab:stage2}
\begin{tabular}{lccc}
\toprule
Family & Solo & Joint $+$ FT & vs.\ solo \\
\midrule
SWE   & $0.0160$ & \best{$0.0014$} & $0.09\times$ \\
DR2D  & $0.1354$ & \best{$0.1203$} & $0.89\times$ \\
NS2D  & $0.4673$ & \best{$0.3788$} & $0.81\times$ \\
CFD2D & $0.1672$ & \best{$0.0887$} & $0.53\times$ \\
Darcy & $0.8904$ & $0.9090$ & $1.02\times$ \\
\bottomrule
\end{tabular}
\end{table}

\paragraph{Five 1D families.} The same recipe with a 1D strip residual also covers
five PDEBench 1D families under a matched solo protocol (per-family nMSE
$0.0001$--$0.076$; see the 1D detail in Appendix~\ref{app:details}), showing the
residual mechanism is not specific to the 2D setting.

\section{Full Cooperation Ablation}
\label{app:ablation}
The complete a--h chain over both fusion modes. Under weighted-sum fusion:
dense soft routing $0.176$, hierarchical $0.276$, cross-expert attention $0.285$, shared
experts $0.326$, gated constraints $0.286$, hard top-1 $0.308$, generator-truth $0.039$.
Under concatenation fusion: dense soft routing $0.137$, hierarchical $0.220$, uniform
$0.283$, hard top-1 $0.301$, generator-truth $0.034$. No cooperation mechanism beyond
dense soft mixing improved the chain in either fusion mode. Concatenation improves
on weighted sum for both dense soft and hierarchical routing, which is why the
primary configuration uses it.

\section{Paired Significance Tests}
\label{app:wilcoxon}
Paired per-trajectory Wilcoxon tests on S2 \R{} ($n{=}384$ held-out
trajectories) rank the primary configuration above every learned baseline at
$p<10^{-40}$. These tests are conditional on the trained models and measure
paired consistency across trajectories; they are not a substitute for
seed-level variation. The only non-significant pair in the full matrix is
GINO versus persistence ($p=0.16$). Seed-level $n{=}3$ statistics are
reported separately and never combined with these.

\section{Compositional Corpus Examples}
\label{app:synthfig}
\begin{figure}[h]
\centering
\includegraphics[width=0.62\textwidth]{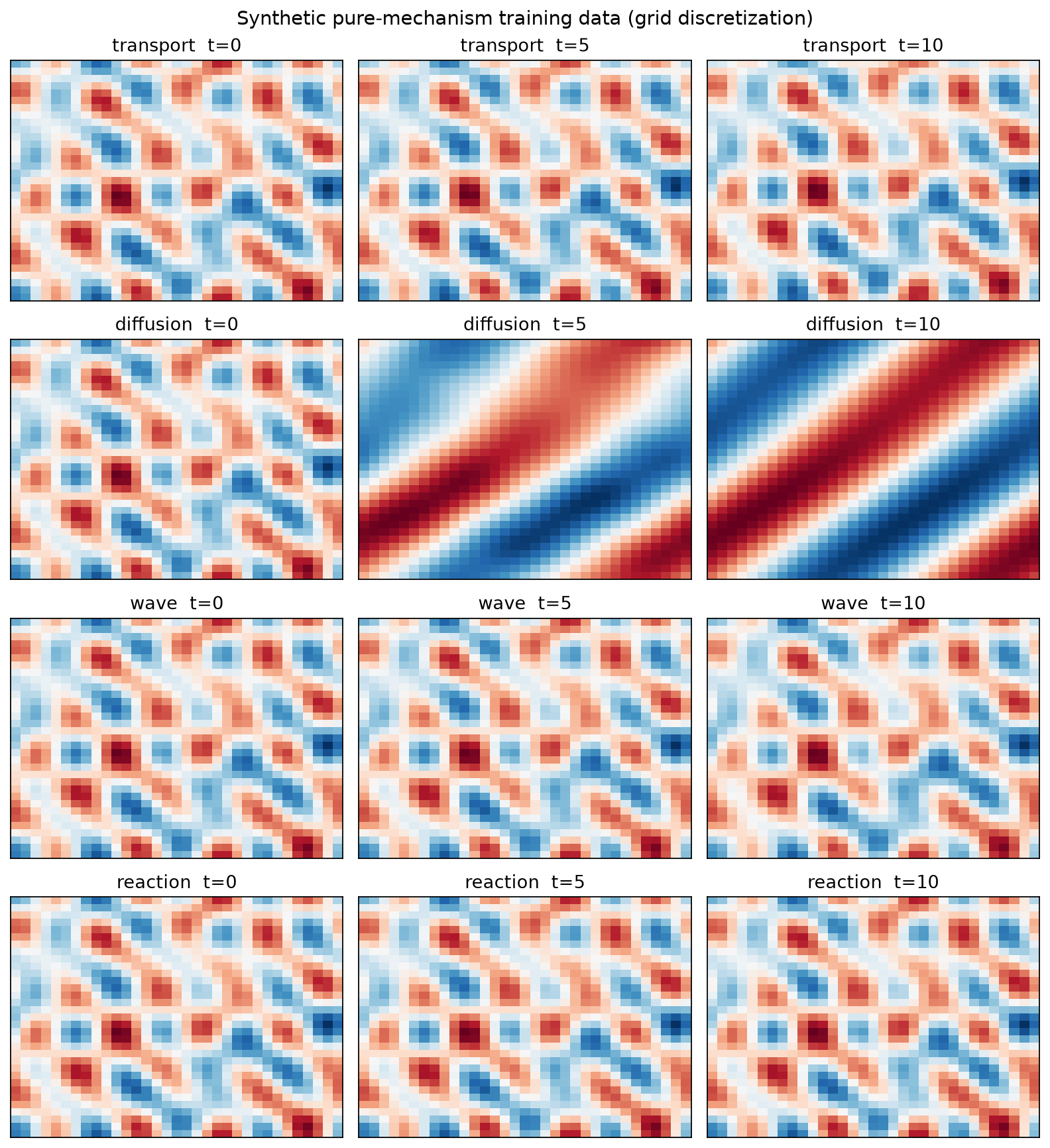}
\caption{Example pure-mechanism training trajectories. Each row is one of the
four generator mechanisms; columns are time snapshots. $w^\star$ is known by
construction for every pure and mixed trajectory because the generator records
the operator-splitting coefficients.}
\label{fig:synthdata}
\end{figure}

\section{Experimental Details}
\label{app:details}

\textbf{Five 1D families.}
\label{sec:1dtable}
Per-family nMSE under the matched solo protocol (3 seeds): advection
$0.041{\pm}0.013$, Burgers $0.038{\pm}0.014$, diffusion--reaction
$0.044{\pm}0.011$, compressible flow $0.076{\pm}0.003$, sorption
$0.0001{\pm}0.000$ (near-floor under this subsample; not over-read).

\textbf{FNO baseline.} 12 retained Fourier modes per dimension, width 32, 4
spectral layers, trained per family with the shared next-step objective and
evaluated under the same rollout harness as COSMOS.

\textbf{Test populations.} Paired per-trajectory tests on the compositional S2
holdout use 384 held-out trajectories; each external PDEBench family is
evaluated on 100 held-out trajectories per seed.

\textbf{Compositional corpus.} On the order of $10^3$ trajectories per family
across four pure and five mixed families, each emitted as grid ($33{\times}33$),
mesh, and point-cloud siblings; splits and generator settings ship with the
corpus.

\textbf{PDEBench sources.} Shallow water uses the 2D radial dam-break subset;
Navier--Stokes uses the incompressible velocity fields; diffusion--reaction uses
the 2D two-channel subset. Up to 400 trajectories per family are adapted to
$64{\times}64$ grids for training and evaluation.

\textbf{Hardware.} Training and evaluation ran on single NVIDIA H200-class
GPUs.

\section{Protocol Seams}
\label{app:seams}
Three groups of numbers appear in this paper and are never averaged across:
(i)~external per-family COSMOS ($\sim$3.4M, $d{=}128$) versus FNO ($\sim$1.2M) on
$64{\times}64$ PDEBench grids, three seeds under the shared evaluation harness;
(ii)~compositional synthetic ($33{\times}33$, $\sim$4M) with mechanism labels;
(iii)~joint pretrain$+$fine-tune ($d{=}256$, $K_m{=}8$, $\sim$16.2M). Every table
is annotated with its group. All headline numbers are drawn from frozen claim
records rather than transcribed from logs.

\end{document}